\documentclass[10pt, a4paper]{article}
 \ifdefined\directlua
  \usepackage{fontspec}
 \else
  \usepackage[T1]{fontenc}
  \usepackage[nomath]{lmodern}
\fi
\usepackage[utf8]{inputenc}
\usepackage{authblk}
\usepackage{ACL2023}

\usepackage{array, multirow}
\usepackage{placeins}
\usepackage{xcolor}
\usepackage{soul}
\usepackage{enumitem}
\usepackage{listings}
\usepackage{hyperref}
\usepackage{amsmath}
\usepackage{ntheorem} 
\theoremstyle{margincmh}
\theorembodyfont{}
\theoremsymbol{}
\theoremprework{\medskip}
\theorempostwork{}
\theoremseparator{:}

\hypersetup{
    colorlinks=true,
    linkcolor=blue,
    urlcolor=blue,
}
\usepackage{graphicx}
\usepackage{svg}
\usepackage{booktabs}
\usepackage{makecell}

\newcommand{\data}{{\sc KGConv}}
\newcommand{\qanlc}{{$D_{QA_{nl}}$}}
\newcommand{\qnlc}{{$D_{Q_{nl}}$}}
\newcommand{\klc}{{$D_{kl}$}}
\newcommand{\hc}{{$D_{QA_{nl}+kl}$}}

\newcommand{\triple}[3]{(\textsl{#1}, \textsl{#2}, \textsl{#3})}

\title{Question Generation in Knowledge-Driven Dialog: Explainability and Evaluation}

\author[*]{Juliette Faille}
\author[**]{Quentin Brabant}
\author[**]{\\ Gwenole Lecorve} 
\author[**]{Lina M. Rojas-Barahona} 
\author[*]{Claire Gardent}

\affil[*]{Université de Lorraine, CNRS/LORIA \texttt{\{juliette.faille, claire.gardent\}@loria.fr}} 
\affil[**]{Orange Innovation \texttt{\{quentin.brabant, gwenole.lecorve, lina.rojas\}@orange.com}}

\begin{document}
\maketitle
\begin{abstract}
We explore question generation in the context of knowledge-grounded  dialogs focusing on explainability and evaluation. 
Inspired by previous work on planning-based summarisation, we present a model which instead of directly generating a question, sequentially predicts first a fact then a question.  
We evaluate our approach on 37k test dialogs adapted from the \data\ dataset and we show that, although more demanding in terms of inference, our approach performs on par with a standard model which solely generates a question while allowing for a  detailed reference-less evaluation of the model behaviour in terms of relevance, factuality and pronominalisation.
\end{abstract}

\section{Introduction}
\label{sec:intro}

Recent years have seen an increasing number of approaches aiming to ground dialog models in general purpose knowledge graphs either to avoid over-generic dialog turns \cite{han-etal-2015-exploiting,sun-etal-2018-open,moon-etal-2019-opendialkg} or to support information seeking dialogs \cite{liu-etal-2019-knowledge,perez-beltrachini-etal-2023-semantic}. Here, we focus instead on how knowledge can be used to support  dialogs where facts are  used to guide the generation of a series of questions intended to quizz a user about a given topic or entity. Specifically, we focus on dialogs where each question-answer pair is grounded in a fact and facts are triples from the Wikidata Knowledge Base (KB). An example dialog is shown in Table~\ref{tab:dialog_example}. 

\begin{table}[ht]
\small
\begin{tabular}{ll}
\toprule

T$_1$ & \triple{Sitara Achakzai}{field of work}{feminism} \\
Q$_1$ & \footnotesize \texttt{What was the field of work of}\\[-1ex]
 &\hspace{3cm}\footnotesize \texttt{Sitara Achakzai?}\\
A$_1$ & \footnotesize \texttt{feminism} \\
   
T$_2$ & \triple{Sitara Achakzai}{death manner}{murder}\\
Q$_2$ & \footnotesize \texttt{What was the cause of death}\\[-1ex]
&\hspace{3cm}\footnotesize \texttt{of Achakzai?} \\

A$_2$ &\footnotesize \texttt{homicide}\\
 
T$_3$ & \triple{Sitara Achakzai}{birthplace}{Afghanistan}\\
Q$_3$ & \footnotesize \texttt{Where was she born ?}\\
A$_3$ & \footnotesize \texttt{Afghanistan}\\
    
T$_4$ & \triple{Afghanistan}{capital}{Kabul}\\
Q$_4$ & \footnotesize \texttt{What is the capital of Afghanistan?}\\
A$_4$ & \footnotesize \texttt{Kabul}\\

T$_5$ &\triple{Afghanistan}{lowest point}{Amu Darya}\\
Q$_5$ & \footnotesize \texttt{What is the lowest point of}\\[-1ex]
&\hspace{3cm}\footnotesize \texttt{Afghanistan?}\\
A$_5$ & \footnotesize \texttt{Amu Darya}\\\bottomrule

\end{tabular}
\caption{Example of dialog in the \data\ dataset (T:Triple, Q:Question, A:Answer).}
\label{tab:dialog_example}
\end{table}

We confront several challenges. First, the generated question must fit the dialog context in that it should be neither redundant nor irrelevant (relevance). Second, it should be factual i.e.,  it should bear on an attested fact (factuality). Third, it should be natural sounding, in particular,  integrating the anaphors that are typical of conversational dialogs (pronouns). 
These three challenges are further compounded by an evaluation challenge: 

\begin{quote}
Given that a dialog can be continued in multiple ways, how can we evaluate the behavior of a dialog model with respect to relevance, factuality and pronominalisation?
\end{quote}
We present an explainable approach to knowledge-grounded, conversational question generation (QG) which permits addressing this question.


In contrast to a standard conversational QG model which generates a question conditioning on some input, we train our question generation model to first generate a KB triple and then continue generating the corresponding question. That is, in our approach, question generation is conditioned on both the input context and a predicted KB triple. This has two key advantages. 

First, it improves explainability\footnote{We adopt  \cite{arrieta_xai_survey}'s definition of explainability: 
a model is explainable  if it can be explained to a human using some kind of interface between the model  and the human. As we shall see, in our case, this interface  is provided by the interplay between dialog context, generated triple and generated question.} as the predicted triple  question generation is conditioned on, can be viewed as a strong indicator of why the model generated that question. 

Second, it helps assessing both factuality  (is the predicted triple a well-formed triple of the KB?) and relevance (is the generated triple in topic and new with respect to the triples already captured by the dialog context?). In addition, comparing the generated question with the predicted triple can be used to  assess the contextual relevance of the generated  question  (does the generated question match the generated triple?) and to verify the correctness of pronouns (does the pronoun genre match the semantic genre of the corresponding entity in the predicted triple and is the pronoun ambiguous given the genre of the previously mentioned entities?) . 

In short,  training a model to generate both a triple and a question instead of only a question enables a fine-grained, automatic and reference-less evaluation of the generated questions. 

We make the following contributions. 

\begin{itemize}
    \item Using the \data\ dataset\footnote{The \data\ dataset and paper are currently under anonymous peer review and the dataset will be made publicly available as soon as the paper is published.}, a dataset of 70K dialogs where each (question,answer) pair is grounded in a Wikidata triple, we demonstrate how our approach permits assessing factuality, relevance and pronouns. 
    \item We compare our model with a standard, question-only generation model and show that while  more demanding in terms of inference and more explainable, our architecture reaches similar overall performance in terms of both automatic (similarity score against mutliple references) and human evaluation scores.
    \item  Using an ablation study, we highlight the impact of knowledge on dialog coherence. Conditioning  dialog response generation on  a knowledge graph (in addition to dialog context) drastically helps reduce the proportion of generated questions that are either off topic or factually incorrect. 
    \item All data, models, and scripts to reproduce our results will be made publicly available.
\end{itemize}



\begin{table}[ht!]
\small
\begin{tabular}{@{}p{1.5cm}@{~~}p{6cm}@{}}
\toprule
$e$ & \scriptsize \texttt{Sitara Achakzai}\\\hline
$T_e$ & \scriptsize \texttt{person} \\\hline
$Len_K$ & \scriptsize \texttt{5} \\\hline
RDF graph $K$ & \scriptsize \texttt{\textbf{<t>} <sj> Afghanistan <p> lowest point <o> Amu Darya \textbf{</t>} \textbf{<t>} <sj> Sitara Achakzai <p> birthplace <o> Afghanistan \textbf{</t>} \textbf{<t>} <sj>Sitara Achakzai <p> field of work <o> feminism \textbf{</t>} \textbf{<t>} <sj> Sitara Achakzai <p> death manner <o> murder \textbf{</t>} \textbf{<t>} <sj> Afghanistan <p> capital <o> Kabul \textbf{</t>} } \\\hline

\qanlc\ & \scriptsize \texttt{\textbf{<q>}What was the field of work of Sitara Achakzai? \textbf{<a>} feminism \textbf{<q>} What was the cause of death of Achakzai? \textbf{<a>} homicide}\\\hline
\qnlc\ & \scriptsize \texttt{\textbf{<q>} What was the field of work of Sitara Achakzai? \textbf{<q>} What was the cause of death of Achakzai? } \\ \hline
\klc & \scriptsize \texttt{\textbf{<t>}  <sj>Sitara Achakzai <p> field of work <o> feminism \textbf{</t>} \textbf{<t>} <sj> Sitara Achakzai <p> death manner <o> murder \textbf{</t>} } \\\hline
\hc & \scriptsize \texttt{\textbf{<t>} <sj>Sitara Achakzai <p> field of work <o> feminism \textbf{<q>}What was the field of work of Sitara Achakzai? \textbf{<a>} feminism \textbf{<t>} <sj> Sitara Achakzai <p> death manner <o> murder \textbf{</t>} \textbf{<q>} What was the cause of death of Achakzai? \textbf{<a>} homicide} \\\hline

Ref & \scriptsize\texttt{[TRIPLE] \textbf{<t>} <sj>Sitara Achakzai <p> birthplace <o> Afghanistan infarction \textbf{</t>} [QUESTION] Where was she born ?}\\
\bottomrule
\end{tabular}
\caption{Input elements and reference output for the generation of the third question of the dialog of Table \ref{tab:dialog_example}}
\label{table:example_input}
\end{table}

\section{Dataset}
\label{sec:data}
 
The \data\ dataset consists of 70,596 
English dialogs such that each dialog is composed of a sequence of question-answer pairs about Wikidata entities. Specifically, each \data\ question-answer pair is grounded in a triple (subject, property, object) whose object is the expected answer. For example, the question-answer pair ``Q: Where was Obama born?, A: Hawaii'' is associated with the triple (Obama, place of birth, Honolulu). 

As illustrated in Table~\ref{table:example_input}, each dialog $D$ in \data\ is associated with a root entity $e$ which is the subject entity of the triple initiating the dialog, the Wikidata category $T_e$ of that entity and the graph $K_D$ consisting of the  Wikidata  triples associated with the question-answer pairs making up $D$.
The number of questions in a dialog is at least 5 and at most 19. 

We write $\mathcal{K}_{KGConv}$ the set of all Wikidata triples (143K triples) grounding the \data{} dialogs ($\mathcal{K}_{KGConv} = \cup_{D\in \data} K_D$).


\section{Problem Formulation}
\label{sec:task}
Given $K = \{f_1, \ldots, f_k\}$, a knowledge graph representing a set of facts, and  $D$ a dialog context,  the task consists in generating an adequate follow-up question $q$  which is grounded in a fact $f_i \in K$. 

To enhance explainability, we propose a model which grounds the generation of the next question $q$, not only in $D$ and $K$, but also in the corresponding fact. 
That is, instead of directly decoding a question $q$ from  $K$ and $D$:

\begin{center}
    $K,D \Rightarrow q$ 
\end{center}

we train an encoder-decoder which decodes the concatenation of $f_q$ and $q$ where $f_q$ is the fact underlying $q$\ :
\begin{center}
    $K,D \Rightarrow f_q, q$
\end{center}

As we shall see in Section~\ref{sec:evaluation}, the underlying motivation for this architecture is that the predicted triple can serve as an interface between the user and the model thereby improving its explainability. 

In what follows, we refer to the two types of models as Question-Only and Extended QG model respectively. 

\section{Input}
Drawing on the hybrid NL/KB nature of the \data\ dataset, we experiment with different knowledge sources and dialog contexts. 

\paragraph{Knowledge Graph.} 
In our KB-based dialog setting, each generated question should be grounded in a fact. This raises the issue of content selection: given some knowledge graph $K$, to what extent does the model learn to select a fact that is both relevant and non redundant with respect to the current dialog context?

To study the ability of our model to select a relevant fact from a Knowledge Graph, we train and test on input  graphs of varying size and relevance. Specifically, we condition generation either on $K_D$, the set of triples associated with dialog $D$ in the \data\ dataset or on  $K^+$, a larger set of triples which includes in addition to $K_D$ ($K_D \subset K^+$), \textit{three} types of "distractor" triples which are not in $K_D$:
\begin{itemize}
\item Out-of-Scope triples (entity):  
triples not in  $\mathcal{K}_{D}$  whose subject is of the same Wikidata category as the root entity of the dialog. 
\item Out-of-Scope triples (property): 
triples  not in $\mathcal{K}_{D}$ whose property appears in $K_D$. 
\item Noise triples: triples that are  not in $\mathcal{K}_{D}$ 
but whose subject, property and object are in  $\mathcal{K}_{KGConv}$.
\end{itemize}

The motivation for these additional triples is that they act as "distractors" for content selection with each distractor type representing a  different type of similarity with  relevant triples (all triples in $K_D$). While Noise triples are markedly distinct from the set $K_D$ of triples associated with the dialog at hand, Out-of-Scope triples are semantically close to triples in $K_D$  and are therefore harder to filter out.


We denote by $K^+_n$ the set $K_D$ with $n$ additional triples of each type (i.e., $K^+_0=K_D$, $K^+_1$ has 3 more triples than $K_D$ and so on). We experiment with $n=0,1,2,3$,
capping the total number of added triples to the size of $K_D$. Thus, there are at most as many noisy and out-of-scope triples as relevant ones. 

\paragraph{Dialog Context.} As explained in Section~\ref{sec:data}, in the  \data\ dataset, each question-answer pair in a dialog
is associated with the corresponding  KG triple e.g.,\\

        \begin{tabular}{l}
        T: \triple{Henri\_Poincare}{birth\_place}{Nancy}\\
        Q: \textit{Where was Henri Poincaré born?}\\
        A: \textit{Nancy}\\\\
        \end{tabular}

We leverage this dual (T, Q:A) information to
compare four ways of representing the context: \qanlc\, the standard
dialog context consisting of all the preceding natural language turns
; \qnlc\, the dialog context consisting of all the preceding natural
language questions; \klc , the factual context consisting of the
sequence of facts grounding the preceding dialog turns; and \hc, a
hybrid context consisting of the sequence of preceding dialog turns
and their associated facts. Table~\ref{table:example_input}
illustrates these four types of contexts.

\section{Experimental Setting}
\label{sec:model_task}

We describe the models and the data used for training and testing.

\subsection{Models}

\paragraph{Extended QG Model.}
We use the pre-trained encoder-decoder \textbf{T5-small}\footnote{
A small model is preferred because we want to study model explainability through factuality, relevance, semantic adequacy and referring expressions.} model and fine-tune it on the \data\ dataset. We use the implementation from Huggingface and train on two GPUs (Nvidia GTX 1080 Ti, 11 GiB) with early stopping criteria on the validation loss. We use greedy decoding.
The input to the model is a concatenation of 5 elements
$$e, T_e, Len_{K}, K, D $$
where $e$ is the root entity of the graph and $T_e$ its Wikidata semantic category. $K$ is a set of RDF triples, $Len_{K}$ is the number of triples in $K$ and $D$ is the dialog context. 
The output of the model is the concatenation of a triple $f$ and a natural language  question $q$, where, as illustrated on the last line of Table~\ref{table:example_input}, $f$ and $q$ are prefixed with the special markers [TRIPLE] and [QUESTION] respectively. At training time, $f$ is the Wikidate triple underlying the natural language question $q$. At test time, it is predicted. 
Note that as we have four context types, we train four different models.

\paragraph{Question-Only QG Model.} As our model is tasked with generating both a triple and a question rather than just a question, it is possible that the added transparency comes at the cost of a degraded performance with respect to a model that directly generates the question. We therefore create a  model with the same input as described above, but whose output is now only the next question in natural language. For this Question-Only model, we use the context type \hc.

\begin{table*}[h]
 \small
\begin{tabular}{p{6cm}rrrrrrrrr}
\toprule
Context Type                                                      & \qanlc &  (\%)    &  \qnlc  &    (\%)    & \klc     &    (\%)    & \hc &    (\%)  \\\toprule

\# test examples                                       & 313583 &      & 321270 &      & 315815 &      & 313865     &      \\
\#  distinct generated triples                           & 16519  &      & 18146  &      & 17875  &      & 16597      &      \\\hline
\textbf{Relevant triples }                                    & 303723 & 96 & 286439 & 89 & 301970 & 96 & 304794      & 97 \\\hline
\multicolumn{1}{r}{Exact match with target}                            & 123684 & 39 & 109031 & 34 & 123605 & 39 & 131453     & 42 \\
\multicolumn{1}{r}{Other triple from input graph ($K_D$}                        & 180039 & 57 & 177408 & 55 & 178365 & 56 & 173341     & 55 \\\hline
\textbf{Irrelevant triples }                                      & 9860   & 3 & 34831  & 11 & 13845  & 4 & 9071       & 3 \\\hline
\multicolumn{1}{r}{Repetitions}                                          & 1788   & 1 & 23149  & 7 & 1308   & 0 & 1705       & 1 \\
\multicolumn{1}{r}{Out-of-scope (entity) triples}             & 305    & 0 & 640    & 0 & 340    & 0 & 398        & 0 \\
\multicolumn{1}{r}{Out-of-scope (property) triples}          & 5327   & 2 & 6987   & 2 & 6437   & 2 & 5448       & 2 \\
\multicolumn{1}{r}{Noise triples}                            &        & 0 & 0      & 0 & 0      & 0 & 0          & 0 \\
\multicolumn{1}{r}{Ill-formed triples}                             & 460    & 0 & 2033   & 1 & 1663   & 1 & 710        & 0 \\
\multicolumn{1}{r}{Triples not in \data\ }                             & 5514   & 2 & 7403   & 2 & 7761   & 2 & 4977       & 2\\
\bottomrule
\end{tabular}
\caption{\textbf{Evaluating the generated triples.} For simplificity we rounded the ratio  - as a result they do not always add up to exactly 100. Note also that, as some categories overlap, the total number of irrelevant triples differ from the sum of all bad cases i.e., Repetitions, OOS (entity), etc.}
\label{table:triple_evaluation}
\end{table*}

\subsection{Data}
\label{subsec:data}
We use the 46,808 dialogs of \data\ training set, 7,477 dialogs of the validation set and 10,753 of the test set. We  augment these sets with corresponding  $K^+_{1}$, $K^+_{2}$ and $K^+_{3}$ Knowledge Graphs as described in Section \ref{sec:task} and obtain 161,142 dialogs for the train set, 25,701 for the validation set and 37,056 for the test set.
To avoid too many repetitions in the training and validation sets, we keep half (randomly selected) of all possible contexts for each dialog. For the test samples, we keep all contexts.
We have 666,711 training instances, 105,665 validation, and 323,826 test instances. We remove instances with lengths longer than the model size, i.e. with more than 512 tokens. This represents for each of the context types less than 2\% of the training samples.



\section{Evaluation methodology and results}
\label{sec:evaluation}

We provide  a quantitative evaluation of our Extended QG model in terms of relevance, factuality and pronominalisation drawing on statistics about the generated triples and the relation between  triples and generated questions. 

Since such statistics are not available for the Question-Only QG model, we carry out a human evaluation to compare their ability to generate relevant queries\footnote{We did not do a human evaluation for factuality as this is much more difficult to assess for humans. } and we use automatic metrics to compare their overall performance.

\subsection{Relevance and Factuality}
To assess whether a question is factual and relevant with respect to a given dialog context, we analyse both the generated triples (Table~\ref{table:triple_evaluation}) and the similarity between predicted triples and the generated questions . 

\paragraph{Triples.} 
We consider a triple to be factual if it is well-formed and present in the \data\ Knowledge-Base\footnote{This is an approximation as the triple could be present in Wikidata or simply denote a true fact  not captured by Wikidata.}.  We consider it to be relevant if it is either an exact match of the reference triple or a triple from the input RDF graph $K_D$ which is not already present in the current dialog context (recall that by construction, triples in $K_D$ are neither out-of-domain nor out-of-scope). 

Table~\ref{table:triple_evaluation} shows detailed triples related statistics for the various types of contexts. We find a low ratio of ill-formed and out of domain triples (Triple not in \data ) which shows that the model learns to mostly predict true facts. 

\begin{table}[h]
\centering
\small
    \begin{tabular}{lllll}
    \toprule
        Context  & \qanlc & \qnlc  & \klc      & \hc \\
        Avg GLEU  & 0.76 & 0.78 & 0.73 &0.76 \\
    \bottomrule
    \end{tabular}
    \caption{Mean of the GLEU scores between a question and the triple it was conditioned on.}
    \label{table:mean_gleu_question_eval}
\end{table}
Similarly, the high ratio of relevant triples (89 to 96\%) indicates a good ability of the model to generate triples that are neither redundant nor irrelevant to the dialog context. The   
\qnlc contexts yield notably more irrelevant  triples (11\%) than the other context types (between 3\% and 4\% of irrelevant triples) and more repeated triples (7\% against 1\% for the other types of context). 
This highlights the fact that to avoid repetitions and irrelevant turns, generation should be conditioned on a dialog context which includes not only questions but also their answers.

Finally, we see that the ratio of generated out-of-scope and noisy triples is low for all models (0 to 2\%) and for all numbers of added distractors triples\footnote{While we do not give the detailed breakdown here because of space restrictions, we found that $K^+_{1}$, $K^+_{2}$ or $K^+_{3}$ yield a similar number of mistakes.}. This indicates that the model can correctly discriminate between relevant and irrelevant triples even when these triples have similarities (shared entity type or property) with relevant ones.


\paragraph{Triple/Question Similarity}
Given that most predicted triples are relevant, if most questions match those predicted triples, we can deduce that most generated questions are likely to  be relevant. To compute the similarity  between  a generated question and verbalizations of the triple it was conditioned on at inference time, we use Google BLEU against multiple references drawing on the fact that, on average, \data\ has  12 references per triple\citep{johnson-etal-2017-googles}.

The results  are shown in Table \ref{table:mean_gleu_question_eval}. We find that the average G-BLEU score is high (from 0.73 to 0.76) across the board. Such high G-BLEU scores  indicate a good match between generated triples and questions suggesting that the generated questions are well-formed relevant questions overall.


\subsection{Pronouns}
\label{subsec:pronouns_eval}
Dialogs typically include pronouns. As in our approach, the generated question is grounded in a fact, we can analyse both whether a pronoun gender matches its referent (the subject entity in the fact the question is conditioned on) and whether its antecedent is easy to identify or in other words, whether the pronoun is ambiguous\footnote{We limit our analysis to third-person pronouns and ignore first and second-person pronouns, which are usually rhetorical. }.
We do this as follows.

\begin{itemize}

\item\textbf{Gender.}
For each entity in our dataset, we retrieve its Wikidata `sex or gender' object if any.  We classify the retrieved genders into three main categories:  feminine, masculine, and other genders (which contain different types of gender queer categories). For the `other genders' category, we accept all pronouns as correct pronouns.
If an entity does not have any `sex or gender' property we consider it as a neutral entity.
To assess whether a pronoun in a generated question has the correct gender, we check whether this gender (either feminine, masculine, neutral, or other gender) is the same as the gender of its referent, i.e. the subject entity of the  triple the question is conditioned on. 

\item \textbf{Ambiguity}
We say that a pronoun with gender $g$ is ambiguous if the last entity of gender $g$ mentioned in the dialog context is not the referent of that pronoun. Figure \ref{fig:example_pronoun_ambiguity} shows  an example. While the referent of the masculine pronoun occurring in the generated question is  "William Hershel", the last entity of masculine gender mentioned in the dialog is "Nevil Maskelyne".
We approximate the number of ambiguous pronouns  using the following two heuristics.  
First, we consider that any pronoun occurring in a null dialog context is ambiguous (since it cannot corefer with a previously mentioned entity).
Second, we count as ambiguous any pronoun in a generated question whose referent is different from the entity of the same gender last mentioned in the dialog.
\end{itemize}


\begin{figure*}[!h]
\small
\begin{tabular}{ll}
\toprule
     Dialog Context & T: \triple{NGC 2539}{discoverer or inventor}{William Herschel} \\
&Q: \footnotesize \texttt{Who found NGC 2423?}\\
&A: \footnotesize \texttt{William Herschel}\\
&\\
&T: \triple{NGC 2539}{constellation}{Puppis}\\
&Q: \footnotesize \texttt{What is the name of the constellation which NGC 2423 belongs?}\\
&A: \footnotesize \texttt{Puppis}\\
&\\
&T: \triple{William Herschel}{student of}{ Nevil Maskelyne})\\
&Q: \footnotesize \texttt{What was the name of Herschel's teacher?}\\
&A: \footnotesize \texttt{Nevil Maskelyne}\\ \hline
     Last feminine entity  & -\\
     Last masculine entity  & \footnotesize \texttt{\textbf{\textcolor{red}{Nevil Maskelyne}}} \\
     Last neutral entity  &\footnotesize \texttt{Puppis} \\
         Generated Question & \footnotesize \texttt{where was \textbf{he} buried?}  \\
     Generated Triple & \triple{William Herschel}{place of burial}{Westminster Abbey} \\
     Pronoun & \footnotesize \texttt{he} \\
     Pronoun Antecedent & \footnotesize \texttt{\textbf{\textcolor{red}{William Herschel}}} \\
     Gender of the pronoun's antecedent & masculine \\
     \bottomrule
\end{tabular}
    \caption{\textbf{Example of Gender Ambiguous Pronoun:} The pronoun denotes a male entity (William Herschel) which is different from the last mentioned male entity (Nevil Maskelyne).}
    \label{fig:example_pronoun_ambiguity}
\end{figure*}


The results of the pronoun evaluation are given in Table \ref {table:results_pronouns_evaluation}. We observe a good ratio of questions containing pronouns (between 8 and 13\% of the test examples) and a good diversity of the triples giving rise to such  questions (about 25\% of the dataset triples lead to the generation of a question containing a pronoun). 
Interestingly, while contexts that contain natural language questions have a similar ratio of questions with pronouns, the  \klc\ context, which only consists of triples, induces a much higher rate of pronouns. 

Focusing on gender first, we observe a strong bias for masculine pronouns and correspondingly a higher error rate ($62\%$) for feminine pronouns when the context is text only. A possible explanation is that, as they are fewer feminine pronouns in the data, the model tends to over-generate masculine pronouns thereby leading to a gender mismatch with the target entity. 
Thus while 
the overall proportion of pronouns with incorrect gender is low (ranging from  3\% to 4\%), this is likely mainly due to a gender bias.  

Regarding ambiguity, we find that the proportion of ambiguous pronouns is quite high, ranging between 29\% and 36\% and suggesting that referring expression generation in dialog is not yet a solved problem. As our counts are based on a heuristics however, a human evaluation would be needed to verify that cases that are deemed ambiguous by these heuristics are not in fact lifted by common sense knowledge. For instance, in the sentence "I hit the window with a stone and it broke", we understand the pronoun "it" as referring to "the window" even though "the window" is not the closest entity of neutral gender.

\begin{table}[h!]
\small
\begin{tabular}{lrrrrr}
\toprule
\multicolumn{2}{l}{Context Type} & \qanlc & \qnlc  & \klc      & \hc \\\hline
\multicolumn{2}{l}{\makecell[l]{questions with\\a pronoun}} & 9\% & 8\% & 13\% & 8\% \\
& ``he''  & 53\% & 47\% & 54\% & 52\% \\
& ``it'' & 32\% &  35\% & 34\%& 35\%\\
& ``him'' &7\% &  10\% &8\% &7\%\\
& ``she'' & 8\% & 7\%  & 3\% & 6\%\\
& ``her''  & <1\% &  1\% & 4\% & <1\% \\\hline
\multicolumn{2}{l}{\makecell[l]{pronouns with\\gender mistakes}} & 5\% & 5\% & 3\% & 4\% \\
& ``he'' &     29\%     &    44\% &   68\%    &   52 \%\%      \\
&``she'' &   62 \%     &    39\%    &  18\% &     34\%     \\
& ``him'' &   4 \%     &   9 \%  &     9  \%   &   8\%    \\
& ``her''  &   3\%    &   5 \%    &     2 \%   &  2\% \\
&  ``it''  &    2\%      &   3\%     &    3 \%  &     4 \%   \\\hline
\multicolumn{2}{l}{\makecell[l]{ambiguous\\pronouns}} & 30\% & 36\% & 34\% & 29\% \\
&  ``it'' &  64\%&  67\% &   76\%   &  66\%\\
&  ``he'' &18\%  & 19\%  &   15\%   &  21\%\\
&``she'' & 14\%&   9\%  &  4\%   & 9\%  \\
&``him'' & 3\% &   4\% &  4\% &   3\%  \\
&``her'' &  1\% & 1\%  &  1\%  & 1\% \\ \hline
\multicolumn{2}{l}{\makecell[l]{pronominalized \\distinct triples}} & 22\% & 19\% & 24\% & 19\% \\\bottomrule
\end{tabular}
\caption{Results of the pronouns evaluation.}
\label{table:results_pronouns_evaluation}
\end{table}



\subsection{Comparison with the Question-Only QG Model}
\label{sec:baseline}

We aim to determine whether the added explainability  supported by our Extended QG model comes at the cost of a performance decline. 
\paragraph{Automatic Evaluation}
Multiple automatic metrics can be used to evaluate the output of a generation model. Since our test set is large (313K instances) and the \data\ dataset makes available a large number of references for each dialog turn (12 on average), we choose to use Google BLEU which is fast to compute  and gives  reasonably reliable results when multiple references are available. 
Specifically, we compute the G-BLEU score between the generated question and the set of references containing the verbalizations of a correct triple (i.e., a triple in the input RDF graph but not in the dialog context). We then select the maximum G-BLEU score found. 

On the test set (313,474 instances), the mean G-BLEU score of the Question-Only QG model is 0,50 against 0.52 for our model. 
The difference is statistically significant (t-test with p $<$ 0.01) suggesting a better overall performance for our model.

\paragraph{Human Evaluation}
We perform a human evaluation on 300 dialog contexts randomly selected  among the person, historical event, food, ideology or country categories, as these are the easiest categories to assess without specific background knowledge. We select 60 dialog contexts of each length from one to five in order to have some diversity in dialog lengths and keep the annotation task reasonably simple. 

Using  the Amazon Mechanical Turk platform, we recruited 5 annotators who have the Amazon Mechanical Turk ``Master Qualification", i.e. who consistently submitted high-quality results in the past. We paid 0.18\$ for dialogs of lengths 1 or 2, 0.27\$ for dialogs of lengths 3 and 4 and 0.63\$ for dialogs of length 5. The total cost of the 1,500 annotations (300 dialogs by five annotators) was 448\$.
 
The annotators were showed with each dialog, the question generated by each of the two models (randomly labeled as questions A and B for the different dialogs).
We ask the 5 annotators to evaluate the question generated by each model along three criteria:
(1)  fluency, (2) whether the generated question is a repetition of information present in the dialog context, and (3) the coherence of each of the generated questions with the dialog context. Fluency and repetition are evaluated on a binary scale (yes, no), coherence on a ternary scale (high, medium, low).
We then use a majority vote among annotators to determine the final annotation of each question and compute the ratio of cases for which each model is assessed to be fluent, non repetitive and coherent. 

We  compute Cohen's kappa and the observed agreement (i.e. the proportion of examples for which annotators agree) between each pair of annotators. We report the mean in Table \ref{tab:iaa}. The relatively low Cohen's kappa and high observed agreement are likely due to the fact that our data is skewed as both the Extended and the Question Only QG model perform on average very well regarding the 3 evaluation criteria.

\begin{table}
    \centering
    \small
   \begin{tabular}{rcc}
   \toprule
         	& \makecell{Cohen's $\kappa$}	& \makecell{Obs. Agr.}\\\hline
fluency (QO)	&0,18	&0,93\\
fluency (M)	&0,19	&0,91\\
repetition (QO)	&0,31	&0,87\\
repetition (M)	&0,26	&0,84\\
coherence (QO)	&0,15	&0,67\\
coherence (M)	&0,16	&0,65\\\bottomrule
    \end{tabular}
    \caption{Evaluation of the inter-annotator agreement for the human evaluation (QO: Question-Only QG model, M: Extended QG model).}
    \label{tab:iaa}
\end{table}

The human evaluation shows similar results for the Question Only QG model and for our model with 93\% of the generated questions judged fluent by the annotators for the Question Only QG model vs. 91\% for our  model. 87\% of the Question Only QG model output was judged non-repetitive for the Question Only QG model vs. 84\% for our model. 67\% of the generated questions was considered coherent with the dialog context for our model against 65\% for the Question Only QG model. Although generating a triple and a question does not seem to improve the quality of the generation, it  does not degrade the quality either. 




\section{Ablation study}
\label{sec:ablation_study}

We use ablation to assess the respective impact of the input knowledge graph and the dialog context on dialog coherence.  We train and test the same model as in Section \ref{sec:model_task} with the exact same data, ablating either the  input knowledge graph or the dialog context from the input.

\paragraph{Ablating the Input Knowledge Graph.}
We hypothesise that by conditioning question generation not only on the dialog context but also on a knowledge graph $K$ helps learn a dialog model which produces coherent sequences of questions. To quantitatively assess the impact of this added input on dialog coherence, we ablate $K_D$ and examine the triples generated by the ablated model. The results are shown in Table~\ref{tab:ablation_rdf_triple_eval}. Depending on the context type between 91\% and 92\% of generated triples are incorrect. Almost all of them (between 81 \% and 84\% of the generated triples) are hallucinated triples not belonging to the set of \data\ triples, a large set of 132K Wikidata triples.
We further investigate for the \hc\ context what these hallucinated triples contain.  In most of the hallucination cases (81\%), even though the triple is not in $\mathcal{K}_{KGConv}$, the subject, the property and the object are in $\mathcal{K}_{KGConv}$. This means that entities and properties from $\mathcal{K}_{KGConv}$ are rearranged into non existing Wikidata triples (examples are given in Figure \ref{fig:ablation_rdf_examples_triples}). 

\begin{figure}[h]
\small
\begin{tabular}{p{7cm}}
\toprule
\textbf{With subject, property and object in $\mathcal{K}_{KGConv}$}\\
(Milky Way, located in the administrative territorial entity, New York City)\\
(Nicolas Louis de Lacaille, place of birth, Paris)\\ \hline

\textbf{\textbf{With object not in $\mathcal{K}_{KGConv}$}} \\
 (Armand David, inv. founded by, Yvonnes-Altas)\\
 (LC3:PE [autophagosome membrane], encoded by, CL3PE)\\
 \bottomrule
\end{tabular}
\caption{Examples of triples generated when ablating RDF graph $K$.}
\label{fig:ablation_rdf_examples_triples}
\end{figure}

\begin{table}[h]
\small
\setlength{\tabcolsep}{2.5pt}
\begin{tabular}{lrrrr}
\toprule
Context Type                                                      & \qanlc   &  \qnlc     & \klc         & \hc   \\\toprule
\# Test examples  & 323k &  302k  & 323k &  323k   \\
Incorrect triple                           &  92\% &  92\% &  91\% &  91\%   \\
Repetition                           &  2\% &  1\% &  2\% &  1\% \\
Triple not in \data\                 &  84\% &  81\% &  83\% &  82\%  \\
Subject not in \data\  &  13\% &  28\% &  17\% &  15\%  \\
Property not in \data\ &  14\% &  33\% &  17\% &  16\%  \\
Object not in \data\  &  13\% &  29\% &  17\% &  15\%\\
\bottomrule
\end{tabular}
\caption{Ablation of RDF graph $K$, Results of triple evaluation.}
\label{tab:ablation_rdf_triple_eval}
\end{table}

\paragraph{Ablating the Context $\mathbf{D}$.}
Unsurprisingly,  ablating the dialog context (Table \ref{table:ablation_context_triple_eval}) drastically reduces the proportion of correct triples (51\%) and increases the ratio of repetitions (46\%). 



Question

\begin{table}[h]
\small
\begin{tabular}{rrr}
\toprule
& \# &\%
\\\midrule
\# test examples                             & 323765 &    \\
Correct triples                           & 166716 & 51   \\
\multicolumn{1}{r}{Exact match with  target}                  & 36474  & 11  \\
\multicolumn{1}{r}{Other triple from input RDF}               & 130242 & 40  \\
Incorrect triples                           & 157049 & 49  \\
\multicolumn{1}{r}{Repetitions }                              & 149363 & 46 \\
\multicolumn{1}{r}{Out-of-scope (entity) triples }   & 327    & 0  \\
\multicolumn{1}{r}{Out-of-scope (property) triples } & 8713   & 3  \\
\multicolumn{1}{r}{Noise triples generated   }                & 0      & 0 \\
\multicolumn{1}{r}{Ill-formed triples }                            & 182    & 0  \\
\multicolumn{1}{r}{Triples with a property not in \data\ }                    & 6989   & 2 \\\bottomrule
\end{tabular}
\caption{Ablation of context $D$, results of triple evaluation.}
\label{table:ablation_context_triple_eval}
\end{table}



\section{Related Work}
\label{sec:related_work}

We review some of the approaches used to evaluate the quality of automatically generated dialogs.



 
\subparagraph{Automatic Dialog Coherence Evaluation.} Various approaches have been proposed to help assess the coherence of a dialog response with the dialog context and/or with a reference turn.  
\citet{dziri-etal-2019-evaluating} measure dialog coherence based on Natural Language Inference (NLI) calculating  whether the dialog context entails, contradicts or is neutral with the generated dialog response.
\citet{lowe-etal-2017-towards} propose the ADEM metric for response evaluation in dialogs which uses both the context and response references. Similarly, 
 the RUBER metric proposed by \citet{tao2017ruber} blends reference and reference-less evaluation approaches, by using evaluating generated turn with respect to both a reference and the input query. More recently, \citep{huang-etal-2020-grade} constructs graphs of the dialog topics  and computes a coherence score based on that graph.

Different from these approaches, we propose a explanation by design architecture which allows for a detailed analysis of the model behaviour. 

\subparagraph{Human Dialog Coherence Evaluation.} Another direction that has been extensively pursued to better analyse the behaviour of dialog models, is the design of human evaluation protocoles. 

For the human evaluation of their knowledge-grounded conversation generation, \citet{zhou-etal-2021-earl} use two criteria, appropriateness and informativeness (which where originally proposed by \citet{10.5555/3304222.3304413}). Appropriateness measures whether the grammar, topic and logic of the response is correct in the conversation context and informativeness whether it mentions some new information compared to the context. In Wizards of Wikipedia, \citet{dinan2019wizard} propose a dialog generation model able to retrieve information from a knowledge base. For its human evaluation they use engagingness which they found to increase when the model provides more new knowledge. Engagingness is not directly and exclusively a measure of coherence but can be expected to drop significantly if the model includes a lot of repetitions or jumps illogically from one topic to another. 
Some papers use a human evaluation criteria which is specific to the coherence of the generated dialog. For instance, \citet{gu-etal-2021-chaincqg} use ``answer consistency'' and ask annotators if the questions and answers are consistent. \citet{shen-etal-2021-gtm} simply use a ``coherence" criterion and ask annotators is the generated question is coherent with the input data (in their case a text).

While we resort to human evaluation to compare our model with a Question-Only model, the core intuition underlying our approach is that evaluation can be facilitated by the model architecture (in our case an alternative decoding strategy), thereby reducing the need for human evaluation. 

\section{Conclusion and Future work}
\label{sec:conclusion}

Evaluation  is a long outstanding issue in dialog modeling. Given the many ways in which a dialog context can be extended, how can we provide reference-less metrics to assess the well-formedness of the model output ? 

In this work, we show that in a conversational setting where each dialog turn is grounded in knowledge, we can draw on the close intertwinning of facts and natural language to design a model which permits such assessment. 
The key underlying idea is that by analysing both the predicted triples and the relation between predicted triple and predicted question, we can have a detailed analysis of  the model behaviour. 

We illustrate the workings of our model  by training it  on the  \data\ dataset, considering  various types of context (questions, question-answer pairs, facts, facts and question-answer pairs), analysing the impact of these contexts on dialog quality and comparing its performance with a question-only generation model. We find that  conditioning generation on both questions and answers or on the corresponding triples is crucial to maintain coherence;  that coreferences are generally better handled when the context includes KB facts; and that pronouns are often ambiguous. We compare our model with a standard, black box, end-to-end QG model and find that while more transparent, our model performs on par with this standard model.  Our ablation study further shows that conditioning question generation on both dialog context and a knowledge graph drastically improves coherence. 


In future work, it would be interesting to investigate how the approach generalises to  knowledge-based dialog datasets such as CSQA \cite{liu-2021-corpus} which includes more complex questions; and to explore ways of mitigating pronoun ambiguity. 


%



\nocite{*}
\section{Bibliographical References}\label{sec:reference}

\bibliography{references}
\bibliographystyle{acl_natbib}




\clearpage




\end{document}